\pdfoutput=1

\documentclass[11pt]{article}

\usepackage[hyphens]{url}  
\usepackage[]{acl}

\usepackage{times}
\usepackage{latexsym}

\usepackage[T1]{fontenc}

\usepackage[utf8]{inputenc}

\usepackage{times}  
\usepackage{helvet} 
\usepackage{courier}  
\usepackage{graphicx} 

\usepackage{natbib}  
\usepackage{xcolor}
\usepackage{caption} 
\usepackage{multirow}
\usepackage{subcaption}
\usepackage{courier}
\usepackage{float}

\title{Heuristic-based Inter-training to Improve Few-shot Multi-perspective Dialog Summarization}

\graphicspath{ {./images/} }

\author{Benjamin Sznajder, Chulaka Gunasekara, Guy Lev,\\ {\bf Eyal Shnarch, Sachindra Joshi, Noam Slonim}
 \\
        IBM Research AI\\
\texttt{\{benjams@il, chulaka.gunasekara@,  guylev@il\}.ibm.com}\\ \texttt{\{eyals@il, jsachind@in, noams@il\}.ibm.com}}
  
\begin{document}
\maketitle
\begin{abstract}
Many organizations require their customer-care agents to manually summarize their conversations with customers. These summaries are vital for decision making purposes of the organizations. The perspective of the summary that is required to be created depends on the application of the summaries. With this work, we study the multi-perspective summarization of customer-care conversations between support agents and customers. We observe that there are different heuristics that are associated with summaries of different perspectives, and explore these heuristics to create weak-labeled data for intermediate training of the models before fine-tuning with scarce human annotated summaries. Most importantly, we show that our approach supports models to generate multi-perspective summaries with a very small amount of annotated data. For example, our approach achieves 94\% of the performance (Rouge-2) of a model trained with the original data, by training only with 7\% of the original data. 

\end{abstract}

\section{Introduction}


Text summarization is the task of creating a short version of a long text, retaining the most important information \cite{Gambhir2016RecentAT}. 
Current work in this area largely focuses on creating summaries for documents ~\cite{Lin2019AbstractiveSA}. Dialog summarization is a variant of text summarization, that aims to summarize two or multi-party dialog while retaining the salient information \cite{goo2018abstractive, shang2018unsupervised, fu-etal-2021-repsum, zou-etal-2021-low}. This occurs frequently in customer-care scenarios \cite{liu2019automatic, zou2020topic, favre-etal-2015-call}, where the customers/users typically call customer-care agents over the phone or chat with support agents using text chat applications in order to resolve product or service issues. Many enterprises require the agents to write a short summary of the dialog for record keeping and decision making purposes. 

\begingroup
\renewcommand{\arraystretch}{0.6} 
\begin{figure}[t]
\centering
\resizebox{0.5\textwidth}{!}{%
\begin{tabular}{p{0.3cm}p{7.5cm}}
\hline
\multicolumn{2}{l}{\footnotesize{Dialog}} \\
\hline
\tiny{\textbf{User}} & \tiny{Hi, I updated my OS a few weeks ago -  since then, I am unable to attach any files into an email. Attachments work with my chat application. Can you please advise?}  \\
\tiny{\textbf{Agent}}  & \tiny{
Thank you for contacting the Help Desk. This issue should be relatively simple to fix.}  \\
\tiny{\textbf{User}}  & \tiny{Thank you}  \\
\tiny{\textbf{Agent}}   &  \tiny{First, please, exit the email app.  Then go into your System Preferences -> Security ; Privacy ;  Security. There, scroll down until you get to Full Disk Access. Unlock your settings (padlock in bottom left corner) and enable the email application there (or add it if it's not listed).
Afterwards, start the App Store, use the search of the Store (top left in the app) and look for "dialog". The only result should be a email app's dialog fix. Run that.  When you start email app next time, you should be able to attach files.} \\
\tiny{\textbf{User}}   &   \tiny{I will try this now}
 \\
\tiny{\textbf{Agent}} & \tiny{Alright, I'll wait here for a confirmation.}
  \\
\tiny{\textbf{User}}  &  \tiny{I'm hitting a problem with my Apple ID -  I get a message to say ``Update Apple ID Settings" -  when I try to enter my password it tells me it's incorrect... I try to reset and I get a screen which says ``An unknown error has occurred"}
  \\
\tiny{\textbf{Agent}}  &  \tiny{I'm afraid there's not much I can do with that -
Apple ID is not in our scope of support :( 
For this 
I'll have to direct you straight to Apple support.}\\
\tiny{\textbf{User}}   &   \tiny{You mean to contact them directly?}\\
\tiny{\textbf{Agent}} & \tiny{Yes -  I can at least find their contact number}\\
\tiny{\textbf{User}} &  \tiny{No don't worry, thank you though. I'll find their website and contact them. When I have the issue fixed I'll open a chat here again}\\
\tiny{\textbf{Agent}} & \tiny{You are very welcome! Would you like my help with anything else?
}\\
\tiny{\textbf{User}} & \tiny{For now I'm good -  thank you \vspace{0.1cm}}\\
\hline
\hline
\multicolumn{2}{l}{\footnotesize{Summary - Customer need}} \\
\hline
& \tiny{The user is unable to attach files into emails. Then the user has an issue with updating Apple ID settings.\vspace{0.1cm}}\\
\hline
\multicolumn{2}{l}{\footnotesize{Summary - Agent answer}} \\
\hline
& \tiny{The agent asked to enable full disk access to the email application and suggested to contact apple support regarding the issue with apple ID. \vspace{0.1cm}}\\
\hline
\multicolumn{2}{l}{\footnotesize{Summary - Customer needs and agent answers (full summary)}} \\
\hline
& \tiny{The customer was unable to attach files to an email after updating his/her OS. The agent guided the customer through the steps to solve the issue. The customer's issue with Apple ID is not in their scope of support.\vspace{0.1cm}}\\
\hline
\end{tabular}
}
\caption{Dialog and its multi-perspective summaries. \vspace{-0.5cm}}
\label{figure:summary-perspectives}
\end{figure}
\endgroup

As multiple stakeholders are involved in a dialog, the intended summary could take many forms depending on the underlying application of the created summaries. Figure~\ref{figure:summary-perspectives} shows an example dialog between a user and a customer-care agent and its summaries from different perspectives. For example, the first summary focuses on the issues raised by the user in the dialog. Such summaries can be utilized in scenarios where the organization is interested in learning about the common problems faced by the users of their services. The second summary focuses on the answers suggested by the agent to resolve the issues, and such summaries can be utilized to learn about the agents behavior in handling user requests. The last summary provides both user and agent perspectives in the form of user request and agent response. Such summaries are important for maintaining customer history records, and can be used to educate new agents to handle similar situations.


Dialog summarization has been studied in the recent years \cite{mccowan2005ami, rameshkumar2020storytelling, gliwa2019samsum, feigenblat2021tweetsumm, chen2021dialsumm}. Although summaries with multiple perspectives are useful for different applications as shown above, all the prior work consider one summary for dialog. Some studies such as \cite{goyal2021hydrasum} focused on creating multiple versions of a summary using many decoders, 
which 
is different from the 
notion of 
multi-perspective summaries. As the summary data collection using crowdsourcing is 
expensive and time-consuming \cite{gunasekara-etal-2021-summary}, amending current datasets with summaries of different perspectives is also an expensive exercise.




In general, there are different heuristics that are associated with summaries of different perspectives. For example, the leading sentences in documents  often contain the most salient information~\cite{zhu2021leveraging}, and 
leveraging this 
has shown good performance in extractive document and conversation summarization~\cite{see-etal-2017-get, gliwa2019samsum, feigenblat2021tweetsumm}. 
Here, we 
observe that in customer-care conversations, the leading utterance made by the customer often relate to the issues 
they face, which is essential 
for summarizing the 
customer's perspective. Similarly, the long sentences in documents typically contain more information compared to short sentences, which is useful for summaries~\cite{gliwa2019samsum, feigenblat2021tweetsumm}. Hence, we 
also observe that in many customer-care conversations, the long utterances 
expressed 
by the agents contain the information needed to resolve user issues, which is useful in agent centric summaries. Similarly, one could 
identify appropriate heuristics for summaries of different perspectives. 

In this work, we investigate the extent by which 
such simple heuristics can be used to generate 
{\it weak-labeled\/} 
summaries for 
dialogs between two or more parties. Then, we add an intermediate training step (inter-training) for dialog summarization models which 
exploits 
these weak-labeled data, before fine-tuning on labeled data. 
We show that 
this inter-training 
enables commonly-used summarization models to fine-tune with a very few examples to generate summaries with different perspectives. In summary, our contributions are as follows; (1) We study the problem of multi-perspective dialog summarization;
to the best of our knowledge, this is the first study to do so; 
(2) We investigate heuristics which can be used to generate weak-labeled data to inter-train dialog summarization models to generate summaries with multiple perspectives; 
and (3) We show that our approach enables the common summarization models to fine-tune with 
only a 
few labeled examples to achieve good accuracy in multi-perspective dialog summarization.
\section{Method} \label{approach}

We 
are interested in multi-perspective dialog summarization between customers and support agents, and we train a model for each perspective. The customer model summarizes the customer's need, raised in the dialog, and the agent model summarizes the answer provided by the agent for that need. Both models start with the same pre-trained model, then each 
is inter-trained with its own weak labeled examples and fine-tuned with the few summaries of its perspective (either customer needs or agent answers). 

We examine two methods for automatically generating weak labeled data for a given side in the dialog; (i) the \textit{Lead} heuristic, in which the first utterance of that side, which contains at least five tokens, is taken as the summary of the given 
side; 
and (ii) the \textit{Long} heuristic, in which the longest utterance of the given side (in terms of number of tokens) is selected as 
the respective 
summary. Both methods can be efficiently used over the entire unlabeled data to extract weak summaries, for each perspective, from (almost) every dialog. 
Clearly, one can come up with other weak labeling heuristics that can be explored within the same paradigm proposed here. 
As this is not the focus of our work, we 
demonstrate 
our method with these simple and 
efficient 
heuristics.

We use the large amount of dialogs with weak summaries, we automatically gathered, to inter-train the pre-trained model. Given a dialog, the model is trained to generate its weak summary. 
Loosely speaking, this aims to improve 
the pre-trained model by bringing it closer to the target task, even 
while using weak and noisy 
supervision. 
By that, presumably, 
the model learns to locate the most important part of an input dialog and to output it as a summary. 
Such model tends towards extractive summarization, due to the nature of the data it was inter-trained 
on. 
This is altered on the next stage, the fine-tuning, where the model learns to generate the desired abstractive, third-person summary. 
Our experiments show that this can be achieved even with only few-shot of fine-tuning. We note that our inter-training approach 
differs from previous work~\cite{zhang2020pegasus, zhu2021leveraging} which mask or remove the target utterance and train the model to generate the missing text.
Specifically, our 
approach yields weak supervision which 
aims to be closer to the target of the downstream task. 
In our experiments we evaluate both approaches.

\section{Experiments and Results} \label{sec:exp}
Next, we validate our claim, that inter-training with automatically generated weak labeled data can improve few-shot multi-perspective dialog summarization.

\begin{figure*}
     \centering
     \begin{subfigure}[b]{0.3\textwidth}
         \centering
         \includegraphics[scale=0.26, trim={1cm 0 2cm 0}, clip]{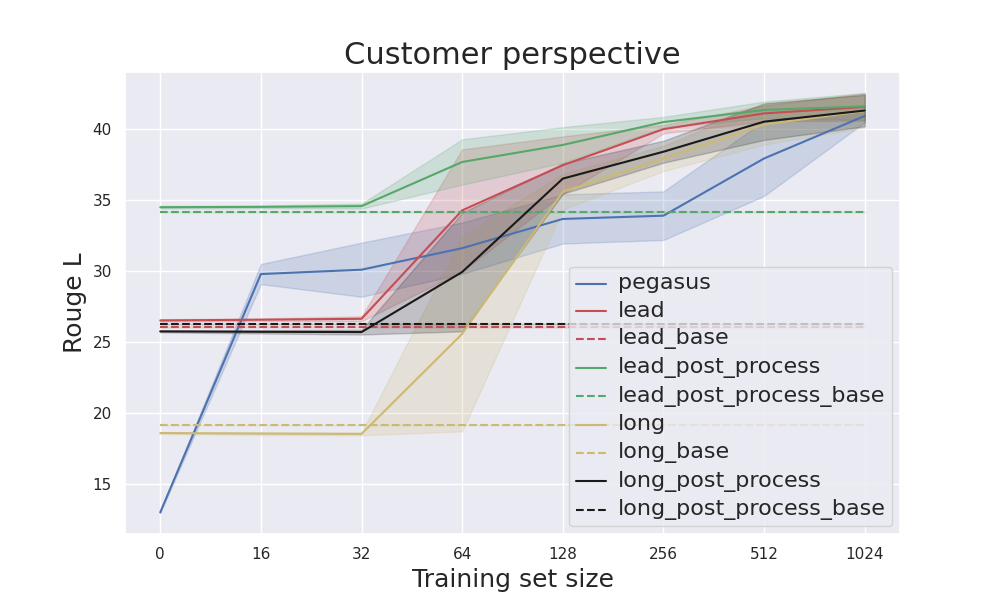}
         \caption{Customer-need perspective}
         \label{fig:y equals x}
     \end{subfigure}
     \hfill
     \begin{subfigure}[b]{0.3\textwidth}
         \centering
         \includegraphics[scale=0.26, trim={1cm 0 2cm 0}, clip]{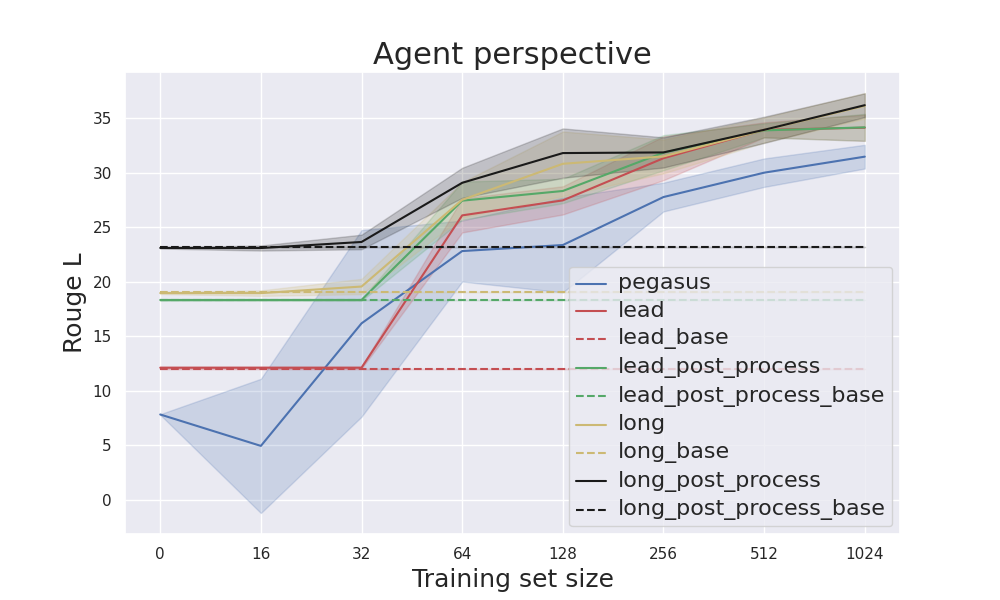}
         \caption{Agent-answer perspective}
         \label{fig:three sin x}
     \end{subfigure}
     \hfill
     \begin{subfigure}[b]{0.3\textwidth}
         \centering
         \includegraphics[scale=0.26, trim={1cm 0 2cm 0}, clip]{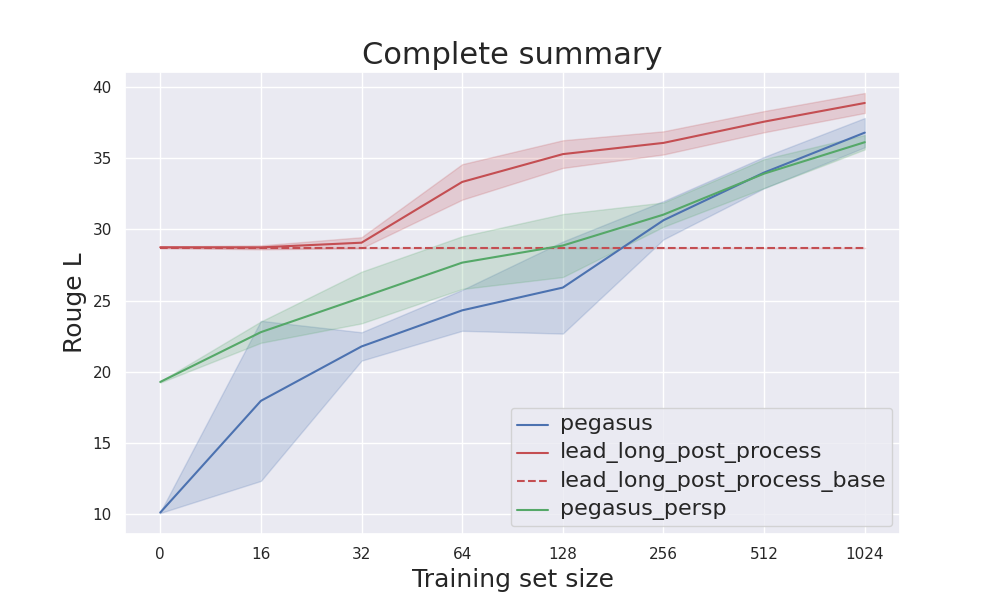}
         \caption{Complete summary}
         \label{fig:five over x}
     \end{subfigure}
        \caption{Rouge-L F-Measure results obtained with different training-set sizes for fine-tuning, for each perspective.}
        \label{fig:three graphs}
\end{figure*}

\subsection{Dataset}
We focus on TweetSumm~\cite{feigenblat2021tweetsumm}, a large scale, high quality, customer-care dialog summarization dataset. TweetSumm consists of 1,100 dialogs, each associated with three extractive and three abstractive summaries. For each dialog in the dataset, we randomly select one abstractive summary as the gold summary.


In TweetSumm, each utterance of a dialog is associated with a speaker (either customer or an agent), and each gold summary consists of two separate fields, where one corresponds to the summary from customer's perspective and the other from agent's perspective. 
Hence, TweetSumm allows easily associating the dialogs and their summaries with the customer's and the agent's perspectives. 
We obtain two views of TweetSumm, each containing all dialogs along with gold summaries limited to one perspective (either the customer or the agent). We use the original split of training (80\%), validation (10\%), and test (10\%) sets of TweetSumm. 



\subsection{Experimental Setup}
We use DistillPegasus \cite{shleifer2020pre} as a pre-trained model in our experiments, but other pre-trained generative models can also be used with our method.
To model the few-shot scenario, we sampled subsets from the training set of sizes $2^n$, where $n\in\{4\cdots 10\}$. Each subset of size $2^i$ is a superset of the smaller subset of size $2^{i-1}$. To this we add the zero-shot scenario, in which no labeled examples are used.
As small training size causes high variance in the results, we repeated these selections of subsets with 5 different random seeds. We report the average performance with its variance over the 5 runs.


As detailed in~\cite{feigenblat2021tweetsumm}, TweetSumm is a subset of a larger dataset of Customer Support dialogs, the \textit{Kaggle Customer Support On Twitter} dataset\footnote{\url{https://www.kaggle.com/thoughtvector/customer-support-on-twitter}}. The weak labeled data is generated only from 22K dialogs in the original larger dataset that are \textit{not} part of TweetSumm. 


As explained in Section~\ref{approach}, we use two heuristics, \textbf{long} and \textbf{lead}, to automatically generate the weak labeled data. We explore additional two variations of these methods:\\
\textbf{long/lead\_masked}: following previous work \cite{zhang2020pegasus, zhu2021leveraging}, we mask the target utterance (longest or lead) from the dialog during inter-training.\\
\textbf{long/lead\_post\_process}: after fine-tuning on the labeled training data, we make a post-processing step to give the generated summaries a more abstractive flavor: we add a prefix of indirect speech clause, in case the summary does not already start with such a clause. Examples of such clauses are ``The customer asks:'' or ``The agent answers:''. We add a prefix to summaries which do not start with ``[The] Customer/Agent''. Figure~\ref{fig:correction} at Appendix \ref{post_proc} depicts the percentage of summaries on which post process was applied, as a function of the training data size. It is noticeable that this percentage drops sharply, with more labeled examples.   

We compare these methods to each other and to several baselines. 
For each of the above methods we have a \textbf{base} version where the obtained summary is the result of applying the heuristic on the test-set with no inter-training (e.g., long\_base for customer perspective outputs the longest customer utterance as the customer side summary for each dialog). In addition, \textbf{pegasus} is a baseline which uses the labeled data for fine-tuning DistillPegasus without the inter-training step.

\subsection{Results}
The evaluation, measured by Rouge-L F-Measure, of the summaries generated by the various inter-trained models is presented in Figure~\ref{fig:three graphs}. Full results, including Rouge-1 and 2, are available in Appendix \ref{more_results}.


We can see in Figure \ref{fig:y equals x}, for the customer perspective, that inter-training with \textit{lead} heuristic outperforms \textit{long}. In Figure \ref{fig:three sin x}, for the agent perspective, we see that inter-training with \textit{long} provides better results compared to \textit{lead}. In both perspectives the post\_process step is very beneficial, and every inter-training method is better than its base version, starting from as little as 32 labeled examples for fine-tuning. 
In addition, in both graphs, the performance of the \textit{pegasus} baseline, where inter-training is not applied, is lower than the inter-training methods for almost all sizes of training sets.
Finally, the results of long/lead\_masked methods (omitted from Figure~\ref{fig:three graphs} for clarity) are worse than the ones obtained by the corresponding long/lead methods when training size is 16 or less. When training size increases, comparable results are obtained.


\subsection{\textit{Divide et Impera} - Back to A Single Summary}
The perspective-based summarization models can be leveraged for creating a classical, full summary of a given dialog. This can be done by simply concatenating the summaries generated by trained customer- and agent-perspective models. We generated full summaries for the test using the best-performing models for each perspective, namely lead\_post\_process for customer and long\_post\_process for agent.


The obtained Rouge-L results are shown by 
lead\_long\_post\_process in Figure~\ref{fig:five over x}. When fine-tuning on the entire training set, 
this approach yields 
a Rouge-L F-measure of 38.89.
It is also important to note that fine-tuning with only \textbf{64} samples, namely \textbf{7\%} of the training set, achieves \textbf{94\%} of the performance of a DistillPegasus model that is not inter-trained, and fine-tuned on the \textbf{entire} original training set (without separation to perspectives). Figure~\ref{fig:five over x} also shows the results of additional baselines. \textit{lead\_long\_post\_process\_base} depicts the concatenation approach when applied with the heuristics themselves: Lead for customer and Long for agent. \textit{pegasus\_persp} depicts the concatenation approach when applied with two DistillPegasus models that are not inter-trained, and fine-tuned on customer and agent perspective each. \textit{pegasus} is a DistillPegasus model fine-tuned with full summaries (without perspective distinction).

The results emphasize the value of our approach: inter-training perspective-focused models by leveraging appropriate heuristics, fine-tuning them (even on very few examples) and combining their outputs leads to significantly better performance.   



\section{Conclusion}



Creating 
gold labeled data for dialog summarization is a demanding process. 
Therefore, many real datasets contain only few conversations along with high quality summaries. In this paper, we investigated how we could utilize 
domain knowledge in this context. 
We proposed the method of inter-training 
on top of weakly labeled summaries 
generated 
based on 
simple heuristics. Our experimental results showed that the inter-training  approach is able to achieve 
excellent 
performance with dramatically less amount of annotated data making it extremely useful in real world settings. Our work 
highlights 
a generic framework that could be used to incorporate heuristics knowledge while training a deep learning model for 
additional tasks. 
In future work, 
we would investigate the potential of additional inter-training heuristics for other types of dialog summarization. 

\nocite{Ando2005,borschinger-johnson-2011-particle,andrew2007scalable,rasooli-tetrault-2015,goodman-etal-2016-noise,harper-2014-learning}

\section*{Acknowledgements}
The authors would like to thank Ranit Aharonov  for her preliminary contribution to this work.

\bibliography{anthology,custom}
\bibliographystyle{acl_natbib}

\appendix

\section{Percentage of Post-Processed Summaries}\label{post_proc}
\begin{figure}[h]
\centering
\includegraphics[scale=0.5]{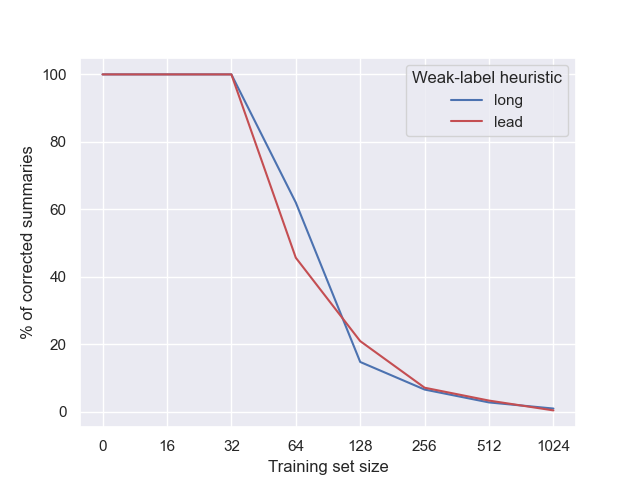}
\caption{Percentage of corrected summaries in the long/lead\_post\_process method.}
\label{fig:correction}
\end{figure}

In Figure \ref{fig:correction} we show how the percentage of post processed summaries changes with increasing training set size of labeled data.

\section{Rouge F-Measure Results Obtained with Different Labeled Training Set Sizes}\label{more_results}

\begin{table*}[t]
\centering
\resizebox{\textwidth}{!}{%
\begin{tabular}{llcccccccc}
\hline
\multicolumn{10}{c}{\textbf{Training set size}}\\ \hline

\textit{\begin{tabular}[c]{@{}l@{}}Rouge\\ Measure\end{tabular}} & \textit{Method Name} & 0 & 16 & 32 &  64 & 128 & 256 & 512 & 1024 \\ \hline
\multicolumn{10}{c}{\textit{Customer perspective}} \\ \hline

\multirow{9}{*}{\begin{tabular}[c]{@{}l@{}}\textit{Rouge-1}
\end{tabular}} 
 & \textit{Pegasus} & $15.12 (\pm0.06)$ &	$31.71 (\pm0.99)$ & $32.09 (\pm2.12)$ & $34.26 (\pm2.09)$ & $36.56 (\pm2.5)$ & $36.49 (\pm1.96)$ & $40.84 (\pm2.82)$ & $43.56 (\pm0.49)$
 \\
 & \textit{Lead} & $29.31 (\pm0.1)$ &	$29.35 (\pm0.11)$ &	$29.44 (\pm0.18)$ &	$37.07 (\pm4.32)$ &	$40.51 (\pm1.97)$ &	$42.65 (\pm0.48)$ &	$44.1 (\pm0.62)$ &	$44.56 (\pm1.22)$  \\
  & \textit{Lead\_base} & $28.71$ & $28.71$ & $28.71$ & $28.71$ & $28.71$ & $28.71$ & $28.71$ & $28.71$ \\
 & \textit{Lead\_post\_process} & $39.19 (\pm0.03)$ &	$39.22 (\pm0.03)$ &	$39.26 (\pm0.08)$ &	$41.25 (\pm1.14)$ &	$42.22 (\pm0.95)$ &	$43.24 (\pm0.6)$ &	$44.31 (\pm0.67)$ &	$44.53 (\pm1.13)$ \\
   & \textit{Lead\_post\_process\_base} & $38.75$ & $38.75$ & $38.75$ & $38.75$  & $38.75$ & $38.75$ & $38.75$ & $38.75$ \\

  & \textit{Long} & $21.91 (\pm0.06)$ &	$21.84 (\pm0.18)$ &	$21.86 (\pm0.15)$ &	$28.65 (\pm6.74)$ &	$38.29 (\pm1.36)$ &	$40.7 (\pm0.7)$ &	$43.12 (\pm1.58)$ &	$44.19 (\pm1.25)$ \\
  & \textit{Long\_base} & $22.28$ & $22.28$ & $22.28$ & $22.28$ & $22.28$ & $22.28$ & $22.28$ & $22.28$\\
 & \textit{Long\_post\_process} & $29.94 (\pm0.07)$ &	$29.89 (\pm0.18)$ &	$29.9 (\pm0.13)$ &	$33.52 (\pm3.74)$ &	$39.35 (\pm1.22)$ &	$41.22 (\pm0.46)$ &	$43.32 (\pm1.36)$ &	$44.24 (\pm1.18)$ \\
  & \textit{Long\_post\_process\_base} & $30.06$ & $30.06$ & $30.06$ & $30.06$ & $30.06$ & $30.06$ & $30.06$ & $30.06$ \\

 \hline
 
\multirow{9}{*}{\begin{tabular}[c]{@{}l@{}}\textit{Rouge-2}
\end{tabular}} 
 & \textit{Pegasus} & $3.65 (\pm0.05)$ &	$14.34 (\pm1.04)$ &	$13.91 (\pm0.9)$ &	$14.14 (\pm0.51)$ &	$15.8 (\pm0.99)$ &	$15.45 (\pm1.04)$ &	$19.07 (\pm2.87)$ &	$22.07 (\pm0.5)$
 \\
 & \textit{Lead} & $14.6 (\pm0.13)$ &	$14.6 (\pm0.13)$ &	$14.62 (\pm0.13)$ &	$19.4 (\pm2.74)$ &	$21.0 (\pm1.15)$ &	$22.18 (\pm0.6)$ &	$22.92 (\pm0.7)$ &	$22.96 (\pm0.54)$  \\
  & \textit{Lead\_base} & $14.32$ & $14.32$ & $14.32$ & $14.32$ & $14.32$ & $14.32$ & $14.32$ & $14.32$ \\
 & \textit{Lead\_post\_process} & $18.27 (\pm0.06)$ &	$18.24 (\pm0.06)$ &	$18.26 (\pm0.12)$ &	$20.75 (\pm1.45)$ &	$21.45 (\pm0.95)$ &	$22.36 (\pm0.65)$ &	$23.04 (\pm0.68)$ &	$23.01 (\pm0.57)$ \\
  & \textit{Lead\_post\_process\_base} & $18.01$ & $18.01$ & $18.01$ & $18.01$ & $18.01$ & $18.01$ & $18.01$ & $18.01$ \\

  & \textit{Long} & $8.38 (\pm0.13)$ &	$8.36 (\pm0.14)$ &	$8.35 (\pm0.18)$ &	$12.7 (\pm4.38)$ &	$19.0 (\pm0.89)$ &	$20.48 (\pm1.21)$ &	$22.06 (\pm1.37)$ &	$22.71 (\pm1.59)$ \\
 & \textit{Long\_base} & $8.61$ & $8.61$ & $8.61$ & $8.61$ & $8.61$ & $8.61$ & $8.61$ & $8.61$\\
 & \textit{Long\_post\_process} & $12.06 (\pm0.07)$ &	$12.04 (\pm0.08)$ &	$12.04 (\pm0.04)$ &	$14.96 (\pm3.01)$ &	$19.37 (\pm0.82)$ &	$20.66 (\pm1.1)$ &	$22.19 (\pm1.24)$ &	$22.77 (\pm1.6)$ \\
 & \textit{Long\_post\_process\_base} & $12.27$ & $12.27$ & $12.27$ & $12.27$ & $12.27$ & $12.27$ & $12.27$ & $12.27$ \\
 
 \hline

\multirow{7}{*}{\begin{tabular}[c]{@{}l@{}}\textit{Rouge-L}
\end{tabular}} 
 & \textit{Pegasus} & $13.03 (\pm0.04)$ &	$29.79 (\pm0.71)$ &	$30.1 (\pm1.9)$ &	$31.61 (\pm1.8)$ &	$33.67 (\pm1.73)$ &	$33.9 (\pm1.71)$ &	$37.92 (\pm2.63)$ &	$40.91 (\pm0.44)$ \\
 & \textit{Lead} & $26.53 (\pm0.08)$ &	$26.58 (\pm0.09)$ &	$26.64 (\pm0.17)$ &	$34.27 (\pm4.29)$ &	$37.45 (\pm2.02)$ &	$39.98 (\pm0.3)$ &	$41.09 (\pm0.55)$ &	$41.54 (\pm0.96)$ \\
 & \textit{Lead\_base} & $26.09$ & $26.09$ & $26.09$ & $26.09$ & $26.09$ & $26.09$ & $26.09$ & $26.09$ \\
 & \textit{Lead\_post\_process} & $34.49 (\pm0.1)$ &	$34.52 (\pm0.09)$ &	$34.57 (\pm0.17)$ &	$37.67 (\pm1.59)$ &	$38.87 (\pm1.25)$ &	$40.48 (\pm0.37)$ &	$41.33 (\pm0.59)$ &	$41.58 (\pm0.94)$ \\
  & \textit{Lead\_post\_process\_base} & $34.16$ &	$34.16$  &	$34.16$  &	$34.16$  &	$34.16$  &	$34.16$  &	$34.16$  &	$34.16$  \\
  & \textit{Long} & $18.61 (\pm0.06)$ &	$18.57 (\pm0.09)$ &	$18.55 (\pm0.1)$ &	$25.59 (\pm6.86)$ &	$35.58 (\pm1.26)$ &	$37.92 (\pm0.89)$ &	$40.31 (\pm1.42)$ &	$41.26 (\pm1.16)$ \\
  & \textit{Long\_base} & $19.17$ & $19.17$ & $19.17$ & $19.17$ & $19.17$ & $19.17$ & $19.17$ & $19.17$\\
 & \textit{Long\_post\_process} & $25.76 (\pm0.07)$ &	$25.73 (\pm0.12)$ &	$25.71 (\pm0.17)$ &	$29.94 (\pm4.17)$ &	$36.5 (\pm1.06)$ &	$38.39 (\pm0.77)$ &	$40.51 (\pm1.28)$ &	$41.29 (\pm1.11)$ \\
  & \textit{Long\_post\_process\_base} & $26.24$ & $26.24$ & $26.24$ & $26.24$ & $26.24$ & $26.24$ & $26.24$ & $26.24$ \\

 \hline
 

 \multicolumn{10}{c}{\textit{Agent perspective}} \\ \hline

\multirow{9}{*}{\begin{tabular}[c]{@{}l@{}}\textit{Rouge-1}
\end{tabular}} 
 & \textit{Pegasus} & $8.92 (\pm0.03)$ & $5.35 (\pm6.7)$ & $18.86 (\pm10.06)$ & $26.28 (\pm2.91)$ & $26.81 (\pm5.23)$ & $31.69 (\pm1.46)$ & $33.53 (\pm1.17)$ & $34.73 (\pm1.32)$
 \\
 & \textit{Lead} & $13.26 (\pm0.1)$ & $13.26 (\pm0.1)$ & $13.26 (\pm0.1)$ & $28.91 (\pm1.67)$ & $30.24 (\pm1.58)$ & $34.46 (\pm2.29)$ & $36.73 (\pm0.53)$ & $37.02 (\pm1.35)$
 \\
 & \textit{Lead\_base} & $13.32$ & $13.32$ & $13.32$ & $13.32$ & $13.32$ & $13.32$ & $13.32$ & $13.32$ \\
 & \textit{Lead\_post\_process} & $19.75 (\pm0.04)$ & $19.75 (\pm0.04)$ & $19.75 (\pm0.04)$ & $30.27 (\pm1.85)$ & $31.21 (\pm1.35)$ & $35.02 (\pm1.83)$ & $36.82 (\pm0.45)$ & $37.17 (\pm1.25)$ \\
   & \textit{Lead\_post\_process\_base} & $19.72$ & $19.72$ & $19.72$ & $19.72$  & $19.72$ & $19.72$ & $19.72$ & $19.72$ \\

  & \textit{Long} & $21.91 (\pm0.06)$ &	$21.84 (\pm0.18)$ &	$21.86 (\pm0.15)$ &	$28.65 (\pm6.74)$ &	$38.29 (\pm1.36)$ &	$40.7 (\pm0.7)$ &	$43.12 (\pm1.58)$ &	$44.19 (\pm1.25)$ \\
 & \textit{Long\_base} & $22.21$ & $22.21$ & $22.21$ & $22.21$ & $22.21$ & $22.21$ & $22.21$ & $22.21$\\
 & \textit{Long\_post\_process} & $26.47 (\pm0.07)$ & $26.45 (\pm0.28)$ & $27.08 (\pm0.79)$ & $33.01 (\pm1.33)$ & $35.51 (\pm2.13)$ & $35.66 (\pm1.41)$ & $37.44 (\pm1.56)$ & $39.59 (\pm1.35)$ \\
  & \textit{Long\_post\_process\_base} &$26.36$ & $26.36$ & $26.36$ & $26.36$ $26.36$ & $26.36$ & $26.36$ & $26.36$ \\

 \hline
 
\multirow{9}{*}{\begin{tabular}[c]{@{}l@{}}\textit{Rouge-2}
\end{tabular}} 
 & \textit{Pegasus} & $1.31 (\pm0.02)$ & $0.39 (\pm0.74)$ & $2.07 (\pm1.61)$ & $4.55 (\pm1.7)$ & $5.06 (\pm2.79)$ & $7.81 (\pm1.18)$ & $10.72 (\pm0.79)$ & $12.47 (\pm1.35)$
 \\
 & \textit{Lead} & $4.75 (\pm0.09)$ & $4.75 (\pm0.09)$ & $4.75 (\pm0.09)$ & $10.21 (\pm1.7)$ & $11.28 (\pm1.34)$ & $14.07 (\pm1.41)$ & $16.39 (\pm0.39)$ & $16.7 (\pm1.11)$  \\
 & \textit{Lead\_base} & $4.70$ & $4.70$ & $4.70$ & $4.70$ & $4.70$ & $4.70$ & $4.70$ & $4.70$ \\
 & \textit{Lead\_post\_process} & $4.85 (\pm0.04)$ & $4.85 (\pm0.04)$ & $4.85 (\pm0.04)$ & $10.23 (\pm1.71)$ & $11.37 (\pm1.38)$ & $14.16 (\pm1.46)$ & $16.56 (\pm0.46)$ & $16.82 (\pm1.1)$ \\
   & \textit{Lead\_post\_process\_base} & $4.80$ & $4.80$ & $4.80$ & $4.80$ & $4.80$ & $4.80$ & $4.80$ & $4.80$ \\

  & \textit{Long} & $9.34 (\pm0.04)$ & $9.33 (\pm0.19)$ & $9.84 (\pm0.56)$ & $13.49 (\pm0.96)$ & $15.52 (\pm1.49)$ & $15.11 (\pm1.45)$ & $17.18 (\pm0.99)$ & $19.26 (\pm0.98)$ \\
 & \textit{Long\_base} & $9.61$ & $9.61$ & $9.61$ & $9.61$ & $9.61$ & $9.61$ & $9.61$ & $9.61$\\
 & \textit{Long\_post\_process} & $9.17 (\pm0.08)$ & $9.15 (\pm0.18)$ & $9.59 (\pm0.46)$ & $13.32 (\pm0.94)$ & $15.4 (\pm1.67)$ & $15.15 (\pm1.44)$ & $17.31 (\pm0.98)$ & $19.31 (\pm0.89)$ \\
 & \textit{Long\_post\_process\_base} & $9.31$ & $9.31$ & $9.31$ & $9.31$ & $9.31$ & $9.31$ & $9.31$ & $9.31$ \\
 
 \hline

\multirow{7}{*}{\begin{tabular}[c]{@{}l@{}}\textit{Rouge-L}
\end{tabular}} 
 & \textit{Pegasus} & $7.81 (\pm0.03)$ & $4.93 (\pm6.16)$ & $16.18 (\pm8.56)$ & $22.81 (\pm2.8)$ & $23.37 (\pm4.35)$ & $27.77 (\pm1.32)$ & $30.01 (\pm1.3)$ & $31.48 (\pm1.09)$ \\
 & \textit{Lead} & $12.11 (\pm0.08)$ & $12.11 (\pm0.08)$ & $12.11 (\pm0.08)$ & $26.09 (\pm1.57)$ & $27.47 (\pm1.29)$ & $31.34 (\pm2.01)$ & $33.94 (\pm0.68)$ & $34.14 (\pm1.21)$ \\
 & \textit{Lead\_base} & $12.01$ & $12.01$ & $12.01$ & $12.01$ \\
 & \textit{Lead\_post\_process} & $18.32 (\pm0.07)$ & $18.33 (\pm0.07)$ & $18.33 (\pm0.07)$ & $27.44 (\pm1.75)$ & $28.33 (\pm1.12)$ & $31.85 (\pm1.61)$ & $33.89 (\pm0.65)$ & $34.17 (\pm1.24)$ \\
& \textit{Lead\_post\_process\_base} & $18.33$ & $18.33$ & $18.33$ & $18.33$ & $18.33$ & $18.33$ & $18.33$ & $18.33$ \\
  & \textit{Long} & $18.96 (\pm0.09)$ & $18.95 (\pm0.25)$ & $19.56 (\pm0.7)$ & $27.52 (\pm1.62)$ & $30.83 (\pm2.96)$ & $31.5 (\pm1.55)$ & $33.93 (\pm1.15)$ & $36.15 (\pm1.17)$ \\
 & \textit{Long\_base} & $19.06$ & $19.06$ & $19.06$ & $19.06$ & $19.06$ & $19.06$ & $19.06$ & $19.06$\\
 & \textit{Long\_post\_process} & $23.1 (\pm0.05)$ & $23.09 (\pm0.23)$ & $23.65 (\pm0.67)$ & $29.08 (\pm1.38)$ & $31.81 (\pm2.26)$ & $31.87 (\pm1.38)$ & $33.94 (\pm1.2)$ & $36.21 (\pm1.09)$ \\
 & \textit{Long\_post\_process\_base} & $23.14$ & $23.14$ & $23.14$ & $23.14$ & $23.14$ & $23.14$ & $23.14$ & $23.14$ \\

 \hline


 \multicolumn{10}{c}{\textit{Complete summary}} \\ \hline

\multirow{3}{*}{\begin{tabular}[c]{@{}l@{}}\textit{Rouge-1}
\end{tabular}} 
 & \textit{Pegasus} & $12.54 (\pm0.03)$ & $20.66 (\pm6.03)$ & $25.66 (\pm1.39)$ & $28.97 (\pm1.99)$ & $31.0 (\pm5.03)$ & $37.3 (\pm1.4)$ & $40.41 (\pm1.05)$ & $43.16 (\pm1.25)$
 \\
 & \textit{Lead\_long\_post\_process} & $36.24 (\pm0.04)$ & $36.28 (\pm0.09)$ & $36.57 (\pm0.42)$ & $40.31 (\pm0.98)$ & $41.69 (\pm1.04)$ & $42.01 (\pm1.02)$ & $43.41 (\pm0.85)$ & $44.72 (\pm1.03)$
 \\
 & \textit{Pegasus\_persp} & $23.92 (\pm0.02)$ & $27.23 (\pm1.51)$ & $31.17 (\pm2.74)$ & $34.52 (\pm2.34)$ & $35.69 (\pm3.54)$ & $37.96 (\pm0.84)$ & $40.37 (\pm1.14)$ & $41.97 (\pm0.58)$ \\
 
 \hline
 
\multirow{3}{*}{\begin{tabular}[c]{@{}l@{}}\textit{Rouge-2}
\end{tabular}} 
 & \textit{Pegasus} & $2.7 (\pm0.04)$ & $6.03 (\pm3.22)$ & $8.15 (\pm0.67)$ & $9.42 (\pm1.02)$ & $10.19 (\pm1.34)$ & $12.62 (\pm1.06)$ & $16.01 (\pm1.26)$ & $19.0 (\pm1.08)$
 \\
 & \textit{Lead\_long\_post\_process} & $14.56 (\pm0.05)$ & $14.54 (\pm0.11)$ & $14.79 (\pm0.28)$ & $17.73 (\pm0.79)$ & $18.96 (\pm0.59)$ & $19.33 (\pm0.99)$ & $20.72 (\pm0.58)$ & $21.98 (\pm0.66)$
 \\
 & \textit{Pegasus\_persp} & $6.13 (\pm0.03)$ & $8.34 (\pm0.58)$ & $8.74 (\pm0.66)$ & $9.85 (\pm1.21)$ & $11.0 (\pm1.88)$ & $12.22 (\pm0.61)$ & $15.56 (\pm1.37)$ & $17.87 (\pm0.68)$ \\
 
 \hline

\multirow{3}{*}{\begin{tabular}[c]{@{}l@{}}\textit{Rouge-L}
\end{tabular}} 
 & \textit{Pegasus} & $10.09 (\pm0.03)$ & $17.94 (\pm5.63)$ & $21.77 (\pm1.0)$ & $24.31 (\pm1.44)$ & $25.91 (\pm3.24)$ & $30.64 (\pm1.35)$ & $34.0 (\pm1.1)$ & $36.8 (\pm1.03)$
 \\
 & \textit{Lead\_long\_post\_process} & $28.75 (\pm0.05)$ & $28.74 (\pm0.14)$ & $29.07 (\pm0.38)$ & $33.34 (\pm1.25)$ & $35.29 (\pm0.97)$ & $36.08 (\pm0.81)$ & $37.58 (\pm0.75)$ & $38.89 (\pm0.7)$
 \\
 & \textit{Pegasus\_persp} & $19.27 (\pm0.05)$ & $22.77 (\pm0.75)$ & $25.21 (\pm1.82)$ & $27.66 (\pm1.85)$ & $28.86 (\pm2.21)$ & $31.04 (\pm0.85)$ & $33.92 (\pm1.02)$ & $36.13 (\pm0.51)$ \\
 
 \hline

\end{tabular}}
\label{tab:ROUGE-F}
\caption{ROUGE F-Measure evaluation for different heuristics}%
\end{table*}

\end{document}